# Automated User Identification from Facial Thermograms with Siamese Networks

Elizaveta Prozorova
*Department of Complex Security of Computing Systems*
*Tomsk State University of Control Systems and Radioelectronics*
Tomsk, Russia
pea@fb.tusur.ru

Anton Konev
*Department of Complex Security of Computing Systems Tomsk State University of Control Systems and Radioelectronics*
Tomsk, Russia

Vladimir Faerman
*Department of Complex Security of Computing SystemsTomsk State University of Control Systems and Radioelectronics*
Tomsk, Russia
fva@fb.tusur.ru

*Abstract*— The article analyzes the use of thermal imaging technologies for biometric identification based on facial thermograms. It presents a comparative analysis of infrared spectral ranges (NIR, SWIR, MWIR, and LWIR). The paper also defines key requirements for thermal cameras used in biometric systems, including sensor resolution, thermal sensitivity, and a frame rate of at least 30 Hz. Siamese neural networks are proposed as an effective approach for automating the identification process. In experiments conducted on a proprietary dataset, the proposed method achieved an accuracy of approximately 80%. The study also examines the potential of hybrid systems that combine visible and infrared spectra to overcome the limitations of individual modalities. The results indicate that thermal imaging is a promising technology for developing reliable security systems.

## I. Introduction

In recent years, biometric authentication has been increasingly adopted in various areas, including security and access control systems. It is based on advanced classification and recognition algorithms that process data acquired from biometric sensors. Common biometric devices include fingerprint scanners, iris scanners, and facial imaging systems [1], [2]. The growing interest in these technologies is driven by their higher reliability and convenience compared to traditional authentication methods such as passwords and physical tokens.

Compared with classical approaches, biometric methods offer several important advantages. One key benefit is the ability to collect biometric data without physical contact between the user and the device, which improves usability and hygiene. Contactless biometric identification systems are widely deployed in banks, airports, government facilities, and other environments with high security requirements [3], [4]. However, traditional biometric methods based on visible-light imaging are subject to significant limitations. These include strong dependence on lighting conditions, sensitivity to changes in facial pose and expression, and vulnerability to spoofing attacks using photographs or masks.

The integration of biometric technologies with artificial intelligence methods enhances security and significantly accelerates the identification process. Neural network–based approaches enable more robust feature extraction and improve recognition accuracy under challenging conditions. In this context, thermal facial imaging in the infrared spectrum is of particular interest, as it reflects individual physiological characteristics of a person. Therefore, biometric authentication supported by intelligent data processing techniques represents a promising direction for the development of modern security systems.

## II. Thermal Imaging Identification

### A. Facial Thermograms as Biometric Identifiers

A thermogram is a thermal image acquired using an infrared camera or a thermal imager. It represents the temperature distribution on the surface of the face. A small sample of raw facial thermograms is shown in Fig. 1 for reference.

This distribution forms a unique thermal pattern that reflects the arrangement of blood vessels beneath the skin. Since the vascular structure is individual for each person, facial thermograms can be considered a reliable biometric identifier [5], [6].

Facial thermograms offer several key advantages as a biometric feature. They are largely independent of a person's visual appearance, remaining unaffected by makeup, aging, plastic surgery, or racial and ethnic characteristics [7], [8], [9]. Identification using thermal patterns is possible under any lighting conditions, from bright daylight to complete darkness, which is crucial for continuous surveillance systems [7], [8], [9]. Thermograms also provide strong resistance to spoofing attacks, as photographs, videos, or masks exhibit fundamentally different thermal profiles compared to a live face, allowing effective detection of impersonation [7], [8], [9]. In addition, thermal imaging is a passive method: the camera does not emit any radiation but only captures the subject's natural heat emission, making the process unobtrusive and non-intrusive for users [7], [8], [9].

Despite these advantages, the method has certain limitations. Thermal profiles are influenced by physiological and external factors, including physical activity, emotional state, food or beverage intake (including alcohol or caffeine), and ambient temperature [9]. Accessories such as glasses can block thermal emission, while breathing causes localized, short-term temperature changes around the nose and mouth. Furthermore, thermal cameras generally have lower spatial resolution than visible-light cameras and tend to be more expensive, which can limit their practical deployment in some applications [9].

*B. Thermal Imaging Spectrum Considerations*

While facial thermograms provide a reliable biometric signature, their effectiveness depends not only on image acquisition and processing but also on the choice of infrared spectral range. Different ranges offer distinct advantages and limitations in capturing thermal information. Understanding these differences is essential for selecting the optimal wavelength band for accurate and robust facial recognition. The following section reviews the main infrared ranges and their relevance to biometric identification.

The infrared spectrum is divided into several ranges, each with unique properties that determine its suitability for biometric systems. Near-infrared (NIR, 0.75–1.4 μm) is moderately applicable, as it primarily records reflected radiation and requires active illumination. It can be useful for compensating for changes in visible lighting but cannot capture the subject's own heat [9], [10], [11]. Short-wave infrared (SWIR, 1.4–3.0 μm) shows promising potential in biometrics, as it also records mainly reflected radiation and can penetrate glass, making it effective for detecting masks or other attempts at disguise. However, it may also require additional illumination [9], [10], [11].

Mid-wave infrared (MWIR, 3.0–8.0 μm) has high applicability because it captures the subject's own thermal emission and is sensitive to hot objects, such as those above 300°C. MWIR sensors are often used for thermal fingerprinting, though they frequently require cooling to maintain performance [9], [10], [11]. Long-wave infrared (LWIR, 8.0–14.0 μm) is considered the most suitable for facial biometric identification. It allows passive thermal imaging, as human body radiation peaks around 37°C, corresponding to wavelengths of 9–10 μm. LWIR cameras, including widely used uncooled microbolometers, provide a strong thermal signal and are well adapted for real-world applications [12], [13].

Very long-wave infrared (VLWIR, >14 μm) is generally not used in biometrics, as it is designed for detecting extremely cold objects in fields like astronomy or remote sensing [14]. As indicated by these considerations, LWIR is the preferred spectral range for facial thermography. Its spectral alignment with human body radiation ensures the strongest signal, while the passive nature of LWIR imaging makes it ideal for unobtrusive and reliable biometric identification [15].

*C. Hardware and Data Considerations*

Recognition accuracy in thermal face identification strongly depends on the characteristics of the thermal camera. Spatial resolution is particularly important. While a minimum resolution of 320×240 pixels can capture basic facial features, reliable recognition generally requires 640×512 pixels or higher. High-resolution images provide more informative data for machine learning algorithms, which is crucial because long-wave infrared (LWIR) images often have lower spatial resolution than visible-light images [16].

Thermal sensitivity, measured as Noise Equivalent Temperature Difference (NETD), is another critical factor. Facial thermal contrasts can be extremely subtle, often just hundredths of a degree, so the camera must reliably detect such small variations. For confident recognition, NETD values should not exceed 30 mK, and for high-precision applications, values of 20 mK or lower are recommended. The LWIR spectral band (8–14 μm) is considered optimal for facial thermography, while a frame rate of at least 30 Hz is necessary to prevent motion blur during natural head movements. Lower frame rates, such as 9 Hz or less, make dynamic identification unreliable.

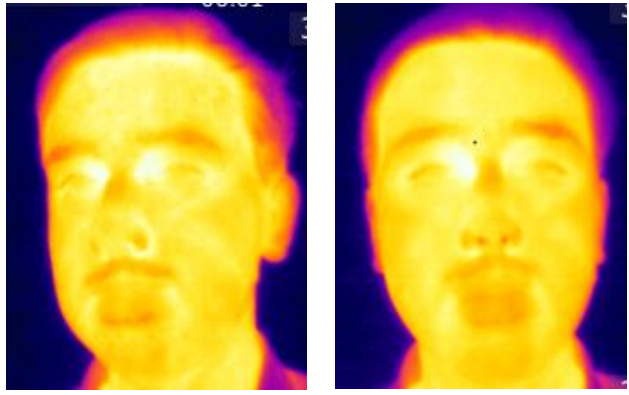

(a)

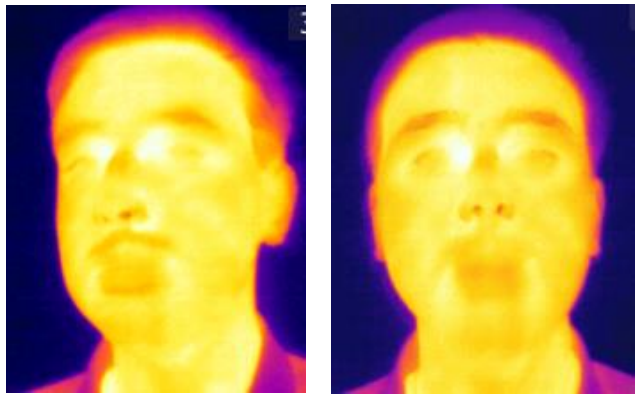

(b)

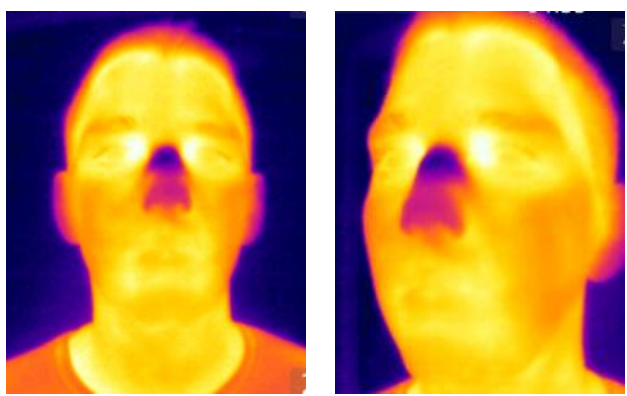

(c)

Fig. 1. Unprocessed facial thermograms: (a) and (b) show the same person imaged under different conditions; (a) and (c) show two different individuals imaged consecutively in a single session. All images are cropped to show only the face region. Images are upscaled for better readability.

Data acquisition mode also affects recognition performance. Thermal information can be captured as static images or video streams, with the latter being preferred in this work. Video allows a single recording to be segmented into thousands of frames, increasing the dataset size and enabling more robust feature extraction. Extracting frames from video also helps reduce overfitting.

When trained on video data, a neural network observes each subject from numerous angles and under varying conditions, including different expressions, partial occlusions, and changes in illumination. This encourages the model to focus on essential and invariant features, such as facial structure and vascular patterns, rather than individual pixels. Consequently, models trained on diverse, high-resolution data are more likely to generalize well to new, unseen real-world scenarios. Modern scientific and industrial thermal cameras, including FLIR A-series, Teledyne FLIR Cortex, Xenics, Lynred, and InfraTec detectors, meet these technical requirements, providing the resolution, sensitivity, spectral range, and frame rate necessary for accurate and reliable facial thermogram recognition.

### D. Security Considerations

Thermal imaging can improve biometric security because it detects heat emitted by the human face, which is harder for simple photos or masks to fake compared to visible-light images. This makes spoofing with common presentation attacks more difficult and increases resistance to basic impersonation attempts in controlled settings [17]. However, thermal systems still face real-world attack risks, as researchers have demonstrated physical perturbations and adversarial patterns that can fool infrared detectors under some conditions, suggesting that robust anti-spoofing defenses are still needed [18].

Despite its advantages, thermal imaging has practical limitations. Thermal images may contain noise and resolution issues that reduce recognition performance, and building large, diverse thermal face datasets for training accurate models remains challenging [19]. Environmental changes and occlusions like eyewear can also affect thermal signatures and lead to errors.

Thermal imaging raises privacy and ethical concerns similar to other biometric technologies. Thermal cameras can reveal personal information beyond identity, such as body shape or health-related data, and their use in public spaces may constitute intrusive monitoring unless properly regulated [18][20]. Therefore, careful design, privacy safeguards, and appropriate legal frameworks are essential when deploying thermal biometric systems in real-world security applications.

## III. PROTOTYPE FOR AUTOMATED IDENTIFICATION

A prototype computer-aided system for facial identification using thermal imaging was developed. It is designed to test the feasibility of thermal biometrics and to evaluate machine learning methods, particularly Siamese neural networks, for comparing facial thermograms. The system combines data acquisition, preprocessing, and feature comparison to enable automated identity verification. The prototype demonstrates how thermal imaging can support secure and reliable biometric recognition in practical applications.

### A. Thermal Images Data Acquisition

The original dataset consisted of 3,720 thermal images. These images were extracted from 32 video streams recorded using a UNI-T UTi260B thermal camera, with each stream lasting between 3 and 5 minutes and corresponding to one of 32 participants. Fig. 2 shows the thermal camera in the experimental setup.

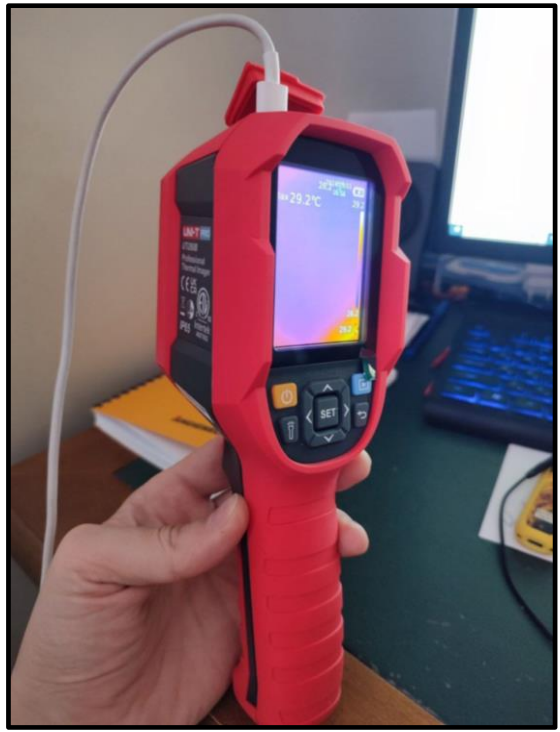

Fig. 2. UNI-T UTi260B thermal camera configured for the automated identification experiment.

Frame extraction from the videos generated the full set of thermograms, providing a rich source of data for training. Using video streams as the source allowed the dataset to capture a wide variety of head angles, facial expressions, and slight movements, which are important for developing robust recognition models. Ambient conditions were kept relatively stable during recording, but natural variations in temperature and posture contributed additional diversity. The resolution and thermal sensitivity of the camera ensured that subtle facial thermal patterns, such as vascular structures, were preserved in the images.

This approach also served as a powerful data augmentation method, increasing the effective size and variability of the dataset without artificial modifications. Each participant contributed multiple frames under slightly different orientations and expressions, helping the model learn invariant features rather than memorizing specific images. This diversity makes the dataset suitable for training Siamese neural networks, which rely on comparing pairs of images to measure similarity. In addition, the dataset reflects real-world conditions, where slight changes in posture, expression, or lighting can occur. Overall, the dataset provides a solid foundation for testing thermal face recognition methods and evaluating their robustness under realistic conditions.

*B. Siamese Neural Network Architecture*

Neural networks are widely used in many applications. This motivated the use of machine learning in this study. In particular, a Siamese neural network was chosen. This architecture has two identical subnetworks that share the same weights [21]. Neural networks are good at automatically extracting complex features from data, such as images or audio. However, standard classifiers are not ideal when the task is to compare two objects and measure how similar or different they are.

The Siamese network works well for comparison tasks. Each subnetwork processes an input image in the same way, extracting important features like the shape of facial regions or vascular patterns. The network converts each image into a feature vector. Vectors of similar images are close to each other, while vectors of different images are far apart. This makes Siamese networks useful for biometric verification, especially when only a few samples per user are available.

In practice, the network takes pairs of images and calculates the Euclidean distance between their feature vectors. A small distance indicates a high chance that the images belong to the same person. A larger distance suggests they belong to different people. Unlike standard classifiers, Siamese networks learn to compare objects rather than recognize specific classes. This approach is ideal for facial thermograms, as it allows accurate identification even with small datasets. Overall, using a Siamese network fits the goal of measuring similarity rather than simple classification, making it a powerful tool for biometric verification.

*C. System Evaluation and Performance*

After preliminary preprocessing, including frame cropping and normalization, the dataset was split into training and test sets in an 80/20 ratio. This setup ensured that the model could learn representative features from the majority of the data while preserving a portion for unbiased evaluation. Each thermal image was paired with another image to form input pairs for the Siamese network, reflecting the task of comparing images rather than performing traditional closed-set classification. This approach aligns with an open-set verification problem, where the model must determine whether two images belong to the same individual, even for identities not seen during training [22].

The Siamese network was trained for 300 epochs using these input pairs. During training, the network learned to encode thermal face images into feature vectors so that similar images were mapped close together and dissimilar images farther apart. Standard data augmentation techniques, such as slight rotations, scaling, and variations in facial expression, were applied to enhance the network's ability to generalize. These practices are common in thermal face recognition and help reduce overfitting, particularly when the dataset is relatively small. Results are presented in the Table I.

TABLE I. RESULTS OF THE IDENTIFICATION PROTOTYPE

| Achieved values (according to common metrics) | | | |
|---|---|---|---|
| *Accuracy* | *Precision* | *Recall* | *F1-Score* |
| 0.7999 | 0.7761 | 0.7899 | 0.7368 |

Model evaluation on the test set demonstrated an accuracy of approximately 80%, with precision, recall, and F1-score indicating balanced performance across the verification task. These results confirm the effectiveness of the Siamese architecture for comparing facial thermograms and highlight its potential for biometric verification in real-world scenarios. Further improvements could be achieved by expanding the dataset, adopting more advanced network architectures such as attention-based models, or fine-tuning hyperparameters to better capture subtle thermal patterns [22], [23].

IV. DISCUSSION

The size of the dataset used in this study, consisting of 3,720 thermal images from 32 participants, is sufficient for developing and testing a prototype system. It allows the network to learn generalizable facial features and to demonstrate the effectiveness of Siamese architectures. However, for a real-world deployment, this dataset is relatively small. Large-scale biometric systems typically require hundreds or thousands of subjects to ensure robustness across diverse populations and environmental conditions.

The identification accuracy of approximately 80% achieved in this study is acceptable for a prototype, confirming the feasibility of the approach. For operational systems in security-critical applications, higher accuracy would be necessary to reduce false acceptances and rejections. This suggests that further improvements, such as increasing dataset size, refining network architecture, or applying advanced augmentation techniques, would be needed for practical implementation.

The use of Siamese networks is particularly suitable for the task, as the system addresses an open-set verification problem where new identities may appear during operation. Traditional closed-set classification would be impractical, because it requires training on all possible identities and cannot generalize to unseen users. The Siamese architecture, by learning a similarity metric instead of class-specific patterns, provides flexibility and allows effective comparison even with limited training samples.

Overall, the results support the potential of thermal facial recognition using Siamese networks as a viable approach for biometric verification. Future research should focus on expanding the dataset, exploring more sophisticated network architectures such as attention-based models, and investigating hybrid systems that combine thermal and visible-light data. Additional studies should also evaluate robustness under varying environmental conditions and across larger, more diverse populations to fully validate the method's practical applicability.

V. CONCLUSION

Facial thermography in the LWIR range (8–14 μm) is a reliable and promising method for biometric identification in high-security applications. It works passively and does not depend on lighting conditions. It is also resistant to spoofing, which makes it suitable for continuous monitoring and access control. Siamese neural networks were used for automated identification. They achieved about 80% accuracy on the collected dataset. Using video streams helped increase data diversity. This allowed the network to learn from different expressions, head positions, and small movements.

Future research should focus on making the system more robust and practical. Possible directions include:

- Creating adaptive algorithms to handle changes in thermal signatures over time;
- Developing hybrid systems that combine visible-light and infrared data;
- Reducing the cost of high-quality LWIR cameras for wider use;
- Combining hardware improvements with smarter algorithms to build reliable, accurate, and user-friendly biometric systems.

ACKNOWLEDGMENT

The authors express their sincere appreciation for the stimulating discussions, as well as the organizational and intellectual support provided through project FEWM-2023-0015 (TUSUR, Ministry of Science and Higher Education of the Russian Federation).